\documentclass{article}
\usepackage{ijcai11}
\usepackage{times}
\usepackage{amsmath}

\usepackage{paralist}

\newtheorem{theorem}{Theorem}
\newtheorem{corollary}%[theorem]
{Corollary}

\newcommand{\systemname}[1]{\emph{#1}}
\newcommand{\cemph}[1]{\textsc{#1}}

\newcommand{\alphabet}{\mathcal{A}}
\newcommand{\calphabet}{\mathcal{C}}
\newcommand{\cassignment}{A}

\newcommand{\head}[1]{head(#1)}
\newcommand{\body}[1]{body(#1)}
\newcommand{\dneg}{not\ }

\newcommand{\domain}[1]{dom(#1)}
\newcommand{\range}[1]{range(#1)}
\newcommand{\scope}[1]{scope(#1)}

\newcommand{\encsup}{$S$}
\newcommand{\encbou}{$B$}
\newcommand{\encran}{$R$}
\newcommand{\encbouh}[1]{\encbou$_{#1}$}
\newcommand{\encranh}[1]{\encran$_{#1}$}

\title{Translation-based Constraint Answer Set Solving}
\author{Christian Drescher \and Toby Walsh \\
NICTA and University of New South Wales, Sydney, Australia}

\begin{document}

% compress equation environments
\abovedisplayshortskip=2pt 
\belowdisplayshortskip=2pt
\abovedisplayskip=3pt
\belowdisplayskip=3pt

\maketitle

\begin{abstract}
We solve constraint satisfaction problems through translation to answer set programming (ASP). Our reformulations have the property that unit-propagation in the ASP solver achieves well defined local consistency properties like arc, bound and range consistency. Experiments demonstrate the
computational value of this approach.
\end{abstract}

\section{Introduction}
%Constraint satisfaction problems (CSP) are combinatorial problems defined as a set of variables whose value must satisfy a number of limitations (the constraints), and stem from a variety of areas in artificial intelligence.
Several formalisms have been proposed for representing and
solving combinatorial problems: constraint programming (CP; \cite{robewa06a}), answer set programming (ASP; \cite{baral03}), propositional satisfiability checking (SAT; \cite{bihemawa09a}), its extension to satisfiability modulo theories (SMT; \cite{niolti06a}), and many more. Each has its particular strengths: for example, CP systems support global constraints, SAT often exploits very efficient implementations, whilst ASP systems permit recursive definitions and offer default negation.
As a non-monotonic reasoning paradigm, ASP is particularly adequate for common-sense reasoning and modelling of dynamic and incomplete knowledge, and was put forward as a powerful paradigm to solve constraint satisfaction problems (CSP) in~\cite{niemela99a}.
Moreover, modern ASP solvers have experienced dramatic improvements in their performance~\cite{gekanesc07a} and compete with the best SAT solvers.
Empirical comparisons with CP have shown that, whilst ASP encodings are often highly competitive and more elaboration tolerant, non-propositional constructs like global constraints are more efficiently handled by CP systems~\cite{dofopo05a}.

This led to the integration of CP with ASP in \emph{hybrid} frameworks, most notably constraint answer set programming (CASP;~\cite{geossc09a}). Similar to SMT, the key idea of a \emph{hybrid} approach is that theory-specific solvers interact in order to compute solutions to the whole constraint model.
However, the elaboration of constraint interdependencies from different solver types is limited by the restricted interface between the ASP and the CP solver.

This paper puts forward a \emph{translation-based} approach rather than a \emph{hybrid} one. In this approach, all parts of the CSP model are mapped into ASP for which highly efficient solvers are available.
We make several contributions to the study of translation into ASP~\cite{drwa10a}:
\begin{compactitem}
\item[-] We consider four different but generic encodings: the direct, support, bound and range encoding. Each represents constraints in a different way.
\item[-] We provide theoretical results on their propagation strength, i.e., what type of local consistency is achieved by the unit-propagation of an ASP solver.
\item[-] We illustrate our approach on the popular \cemph{all-different} constraint. This ensures that a set of variables take all different values. Unit-propagation on our encodings can simulate complex propagation algorithms with a similar overall runtime complexity.
\item[-] We conduct experiments on CSPLib~\cite{gewa99a}, a large problem library widely used for benchmarking by the CP community. Our results demonstrate the competitiveness of this approach.
\end{compactitem}

\section{Background}

\paragraph{Answer Set Programming}
As a form of logic programming oriented towards solving CSP, ASP comes with an expressive but simple modelling language.
Formally, a \emph{logic program} over a set of primitive propositions $\alphabet$, $\bot \in \alphabet$, is a finite set of \emph{rules} $r$ of the form
\[
h \leftarrow a_1 , \dots , a_m, \dneg a_{m+1} , \dots , \dneg a_n
\]
where $h, a_i \in \alphabet$ are \emph{atoms}, $1 \leq i \leq n$. A \emph{literal} is an atom $a$ or its default negation $\dneg a$. The special atom $\bot$ denotes a proposition that is always false. For a rule~$r$, define $\head{r} = h$ and $\body{r} = \{a_1 , \dots , a_m, \dneg a_{m+1} , \dots , \dneg a_n\}$. Furthermore, let $\body{r}^{+} = \{a_1 , \dots , a_m\}$ and $\body{r}^{-} = \{a_{m+1} , \dots , a_n\}$. A rule $r$ with $\head{r} = \bot$ is widely referred to as an \emph{integrity constraint}.
The semantics of a logic program is given by its answer sets, which are the key objects of interest in this paradigm. Given a logic program~$P$ over $\alphabet$, a set~$X \subseteq \alphabet$ is an \emph{answer set} of $P$ iff $X$ is the $\subseteq$-minimal model of the \emph{reduct}~\cite{gellif88b}
\[
P^{X} = \{ \head{r} \leftarrow \body{r}^{+} \mid r \in P,\ \body{r}^{-} \cap X = \emptyset\}.
\]
Intuitively, a rule~$r$ of the form above can be seen as a condition on the answer sets of a logic program, stating that if $a_1 , \dots , a_m$ are in the answer set and none of $a_{m+1} , \dots , a_n$ is included, then $h$ must be in the set.
We also consider extensions to logic programs, such as choice rules and cardinality rules.
A \emph{choice rule} of the form
\[
\{h_1, \dots, h_k\} \leftarrow a_1 , \dots , a_m, \dneg a_{m+1} , \dots , \dneg a_n
\]
allows for the nondeterministic choice over atoms in $\{h_1, \dots, h_k\}$.
A~\emph{cardinality rule} of the form
\[
h \leftarrow k \{a_1 , \dots , a_m, \dneg a_{m+1} , \dots , \dneg a_n\}
\]
infers $h$ if $k$ or more literals in the set $\{a_1 , \dots , a_m$, $\dneg a_{m+1} , \dots , \dneg a_n\}$ are satisfied.
The semantics of choice rules and cardinality rules is given through program transformations (cf. \cite{siniso02a}). % However, modern ASP solvers also incorporate specialised propagators.
Note that aggregations and other forms of set constructions are also common in ASP. However, we will limit ourselves to the above concepts as they are expressive enough for what follows.
Also note that, although answer set semantics is propositional, atoms in~$\alphabet$ can be constructed from a first-order signature. The logic program over $\alphabet$ is then obtained by a \emph{grounding} process, systematically substituting all occurrences of first-order variables with terms formed by function symbols and constants given through the signature.
The task of ASP systems is to compute answer sets for logic programs. A successful framework is conflict-driven nogood learning~(CDNL;\cite{gekanesc07a}). It reflects conditions from program rules in a set of \emph{nogoods}, and describes ASP inference as unit-propagation on nogoods to determine logical consequences.

\paragraph{Constraint Satisfaction and Consistency}
We want to use ASP to model and solve CSP.
Formally, a CSP is a triple $(V,D,C)$ where $V$ is a finite set of \emph{variables}, each $v \in V$ has an associated finite \emph{domain} $\domain{v} \in D$, and $C$ is a set of constraints. A \emph{constraint}~$c$ is a pair~$(R_S,S)$ where $R_S$ is a $k$-ary relation, denoted $\range{c}$, on the variables in $S \in V^k$, denoted $\scope{c}$.
Given a \emph{(constraint variable) assignment} $\cassignment : V \to \bigcup_{v \in V} \domain{v}$, for a constraint~$c$ with $\scope{c} = S = (v_1, \dots, v_k)$ define $\cassignment(S) = (\cassignment(v_1), \dots, \cassignment(v_k))$ and call $c$ \emph{satisfied} if $\cassignment(S) \in \range{c}$.
Define the set of constraints satisfied by $\cassignment$ as $sat_C(\cassignment) = \{ c \mid \cassignment(\scope{c}) \in \range{c},\ c \in C\}$.
A \emph{binary} constraint~$c$ has $|\scope{c}|=2$. For instance, the constraint $v_1 \neq v_2$ ensures that $v_1$ and $v_2$ take different values. An $n$-ary constraint~$c$ has parametrised scope. For instance, \cemph{all-different} ensures that a set of variables, $|\scope{c}|=n$, take all different values. As any non-binary constraint, this can be \emph{decomposed} into binary constraints, i.e., $O(n^2)$~constraints $v_i \neq v_j$ for $i<j$. However, as we shall see in the following, such reformulation can hinder inference.

An assignment~$\cassignment$ is a \emph{solution} to a CSP iff it satisfies all constraints in $C$.
Typically, CP systems use backtracking search to explore assignments in a search tree. In a search tree, each node represents an assignment to some variables, child nodes are obtained by selecting an unassigned variable and having a child node for each possible value for this variable, and the root node is empty. Every time a variable is assigned a value, \emph{constraint propagation} is executed, pruning the set of values for the other variables, i.e., enforcing a certain type of local consistency such as arc, bound, range, or domain consistency.
A binary constraint~$c$ is \emph{arc consistent} iff a variable~$v_1 \in \scope{c}$ is assigned any value~$d_1 \in \domain{v_1}$, there exists a compatible value~$d_2 \in \domain{v_2}$ for the other variable~$v_2$.
An $n$-ary constraint~$c$ is \emph{domain consistent} iff a variable~$v_i \in \scope{c} = \{v_1, \dots, v_n\}$ is assigned any value~$d_i \in \domain{v_i}$, there exist compatible values in the domains of all the other variables~$d_j \in \domain{v_j}$, $1 \leq j \leq n,\ j \neq i$.
Bound and range consistency are defined for constraints over finite intervals. A constraint~$c$ is \emph{bound consistent} iff a variable~$v_i$ is assigned $d_i \in \{min(\domain{v_i}), max(\domain{v_i})\}$ there exist consistent values between the minimum and maximum domain value for all the other variables in the scope of the constraint, called a \emph{bound support}. A constraint is \emph{range consistent} iff a variable is assigned any value in its domain, there exists a bound support. Range consistency is in between domain and bound consistency, where domain consistency is the strongest of the three local consistency properties.

\paragraph{Constraint Answer Set Programming}
Constraint logic programming naturally merges CP and logic programming, while preserving the advantages of either approach to modelling and solving CSP.
Formally, a \emph{constraint logic program} is a logic program~$P$ over an alphabet distinguishing regular atoms~$\alphabet$ and constraint atoms~$\calphabet$, such that $\head{r} \in \alphabet$ for each $r \in P$~\cite{geossc09a}.
A function $\gamma : \calphabet \to C$ associates constraint atoms with constraints. (The set~$C$ stems from the definition of CSP.) For sets of constraints $C' \subseteq C$ define $\gamma(C') = \{ \gamma(c) \mid c \in C' \}$.
Given a constraint logic program~$P$ over $\alphabet$ and $\calphabet$, and an assignment~$\cassignment$, a set~$X \subseteq \alphabet$ is a \emph{constraint answer set} of $P$ with respect to $\cassignment$ iff $X$ is an answer set of the \emph{constraint reduct}~\cite{geossc09a}:
\[
\begin{array}{l}
P^{\cassignment} = \{ \head{r} \leftarrow \body{r}|_\alphabet \mid r \in P,\\
\hspace{3.5cm}\gamma(\body{r}^{+}|_\calphabet) \subseteq sat_C(\cassignment),\\
\hspace{3.5cm}\gamma(\body{r}^{-}|_\calphabet) \cap sat_C(\cassignment) = \emptyset\}.
\end{array}
\]
The idea in our translation-based approach to constraint answer set solving is to compile a constraint logic program into a (normal) logic program by adding an ASP reformulation of constraint variables and  all constraints that appear in the constraint logic program. This allows us to apply CDNL to compute constraint answer sets. A key advantage is that nogood learning techniques can exploit constraint interdependencies since all variables will be shared between constraints. This can improve propagation between constraints. Our reformulations also provides a propagator for the negation of a constraint.

\section{Reformulating CASP into ASP}
We now present four ASP encodings for variables and constraints over finite domains.
All constraints~$c$ are reified via atoms~$sat(c)$, and $violate(c)$, indicating whether~$c$ is satisfied or violated, respectively. To ensure consistency, i.e., either $sat(c)$ or $violate(c)$ is in an answer set, we post
\[
\begin{array}{ll}
sat(c) \leftarrow \dneg violate(c) \\
violate(c) \leftarrow \dneg sat(c)
\end{array}
\]
for every constraint~$c$. % in a given constraint logic program~$P$.
Other representations, e.g., using choice rules, are also possible.
To save the reader from multiple superscripts, in the following, we will assume $\domain{v} = \lbrack 1, d\rbrack$ for all $v \in V$.

\paragraph{Direct Encoding}
A straightforward encoding is the \emph{direct encoding} in which an atom~$e(v,i)$ is introduced for each constraint variable~$v$ and each value~$i$ from their domain, representing~$v = i$. Intuitively, $e(v, i)$ is in an answer set if $v$ takes the value~$i$, and it is not if $v$ takes a value different from~$i$. For each~$v$, possible assignments are encoded by a choice rule (\ref{direct:1}).
Furthermore, we specify that $v$ takes at least one value (\ref{direct:2}) and that it takes at most one value (\ref{direct:3}).
\begin{align}
\{ e(v, 1), \dots, e(v, d) \} &\leftarrow \label{direct:1}\\
\bot &\leftarrow \dneg e(v, 1), \dots, \dneg e(v, d) \label{direct:2}\\
\bot &\leftarrow 2\ \{ e(v, 1), \dots, e(v, d) \} \label{direct:3}
\end{align}
A constraint $c$ is encoded as forbidden combination of values, i.e., if $v_1 = d_1$, $v_2 = d_2$, $\dots$, $v_n = d_n$ is such a forbidden combination then we encode
\[
violate(c) \leftarrow e(v_1, d_1), e(v_2, d_2), \dots, e(v_n, d_n).
\]
%On the other hand, when a relation is represented by allowed combinations of values, one either deduces all forbidden combinations or encode the allowed combinations, conversely, as below
%\[
%sat(c) \leftarrow e(v_1, d_1), e(v_2, d_2), \dots, e(v_n, d_n).
%\]
Unfortunately, the direct encoding hinders propagation:
\begin{theorem}
Enforcing arc consistency on the binary decomposition of a constraint prunes more values from the variables domain than unit-propagation on its direct encoding.
\end{theorem}
The support encoding has been proposed in the domain of SAT to tackle this weakness~\cite{gent02}.

\paragraph{Support Encoding}
We now encode support information for assignments rather than the encoding of conflicts. For each possible assignment to a variable one of its supports must hold, that is, the set of values for the other variable which allow this assignment.
Formally, a \emph{support} for a constraint variable~$v$ to take the value~$i$ across a constraint~$c$ is the set of values $\{i_1, \dots, i_m\} \subseteq \domain{v'}$ of another variable in~$v' \in \scope{c}\setminus \{v\}$ which allow $v = i$, and can be encoded in the following rule, based on (1--3):
\[
violate(c) \leftarrow e(v, i), \dneg e(v', i_1), \dots, \dneg e(v', i_m).
\]
It can be read as whenever $v = i$, then at least one of its supports must hold, otherwise the constraint is violated.
In the support encoding, for each constraint~$c$ there is one support for each pair of distinct variables $v, v' \in \scope{c}$, and for each value $i$.
\begin{theorem}
Unit-propagation on the support encoding enforces arc consistency on the binary decomposition of the original constraint.
\end{theorem}
We have used program transformation~\cite{siniso02a} in~\cite{drwa10a} to reformulate \cemph{all-different} straightforwardly according to our support encoding into $\mathcal{O}(d)$ cardinality rules:
\begin{align}
violate(c) &\leftarrow 2\ \{ e(v_1, i), \dots, e(v_n, i) \}
\end{align}
\begin{corollary}
Unit-propagation on (1--4) enforces arc consistency on the binary decomposition of \cemph{all-different} in $\mathcal{O}(nd^2)$ down any branch of the search tree.
\end{corollary}
%
% other encodings
% His encoding has been generalised by \citet{behewa03a} to $n$-ary CSP, showing that, then, unit-propagation establishes relational $k$-arc consistency on the original problem. \cite{drwa10b}

\paragraph{Range Encoding}
In the \emph{range encoding}, we represent that a variable can take values from an interval $v \in [l,u]$, i.e., a value between $l$ and $u$ (inclusive). An atom~$r(v, l, u)$ is introduced for each $v$ and $[l, u] \subseteq [1, d]$. For each range~$\lbrack l, u \rbrack$, the following $\mathcal{O}(nd^2)$ rules encode $v \in \lbrack l , u \rbrack$ whenever $v \not\in \lbrack 1, l-1 \rbrack$ and $v \not\in \lbrack u+1, d\rbrack$, and enforce a consistent set of ranges, i.e., $v \in \lbrack l, u\rbrack$ implies $v \in \lbrack l-1, u\rbrack$ and $v \in \lbrack l, u+1\rbrack$:
\begin{align}
r(v, l, u) &\leftarrow \dneg r(v, 1, l-1), \dneg r(v, u+1, d) \\
\bot&\leftarrow r(v, l-1, u), \dneg r(v, l, u) \\
\bot&\leftarrow r(v, l, u+1), \dneg r(v, l, u)
\end{align}
Constraints are encoded into integrity constraints representing conflict regions $v_1 \in \lbrack l_1, u_1\rbrack, \dots, v_n \in \lbrack l_n, u_n\rbrack$:
\[
violate(c) \leftarrow r(v_1, l_1, u_1), \dots, r(v_n, l_n, u_n)
\]
\begin{theorem}
Unit-propagation on the range encoding enforces range consistency on the original constraint.
\end{theorem}
An efficient propagator for \cemph{all-different} enforces range consistency by pruning Hall intervals~\cite{le96a}. A Hall interval of size~$k$ completely contains the domains of $k$~variables, formally, $|\{ v \mid dom(v) \subseteq \lbrack l, u \rbrack \}| = u - l + 1$. Observe that in any bound support, the variables whose domains are contained in the Hall interval consume all values within the Hall interval, whilst any other variable must find their support outside the Hall interval (cf. \cite{bekanaquwa09a}).
We encode \cemph{all-different} such that no interval $\lbrack l, u\rbrack$ can contain more variables than its size:
\begin{align}
violate(c)&\leftarrow u-l+2\ \{ r(v_1, l, u), \dots, r(v_n, l, u) \}. \label{range:all}
\end{align}
This simple reformulation can simulate a complex propagation algorithm like the one in~\cite{le96a} with a similar overall complexity.
\begin{corollary}
Unit-propagation on (5--8) enforces range consistency on \cemph{all-different} in $\mathcal{O}(nd^3)$ down any branch of the search tree.
\end{corollary}

\paragraph{Bound Encoding}
In our \emph{bound encoding}, similar to the \emph{order encoding}~\cite{tatakiba06a}, an atom~$b(v, i)$ is introduced for each variable~$v$ and value~$i$ to represent that~$v$ is bounded by~$i$, i.e., $v \leq i$. For each $v$, possible assignments are encoded by a choice rule (\ref{bound:1}). To ensure a consistent set of bounds, (\ref{bound:2}) encodes that $v \leq i$ implies $v \leq i+1$. Finally, (\ref{bound:3}) encodes $v \leq d$, i.e., some value must be assigned to~$v$.
\begin{align}
\{ b(v, 1), \dots, b(v, d) \} &\leftarrow \label{bound:1}\\
\bot&\leftarrow  b(v, i), \dneg b(v, i+1) \label{bound:2}\\
\bot&\leftarrow \dneg b(v, d) \label{bound:3}
\end{align}
Similar to the range encoding, we represent conflict regions $l_1 < v_1 \leq u_1$, $\dots$, $l_n < v_n \leq u_n$ as below
\[
\begin{array}{ll}
violate(c) \leftarrow\!\!\!\!& b(v_1, u_1), \dots, b(v_n, u_n),\\ &\dneg b(v_1, l_1), \dots, \dneg b(v_n, l_n).
\end{array}
\]
\begin{theorem}
Unit-propagation on the bound encoding enforces bound consistency on the original constraint.
\end{theorem}
In order to achieve a reformulation of \cemph{all-different} that can only prune bounds, the bound encoding for variables is linked to (\ref{range:all}) as follows:
\begin{align}
r(v, l, u) &\leftarrow \dneg b(v, l-1), b(v, u) \\
\bot&\leftarrow r(v, l, u), b(v, l-1) \\
\bot&\leftarrow r(v, l, u), \dneg b(v, u)
\end{align}
\begin{corollary}
Unit-propagation on (8--14) enforces bound consistency on \cemph{all-different} in $\mathcal{O}(nd^2)$ down any branch of the search tree.
\end{corollary}

\section{Experiments}
We have conducted experiments on hard combinatorial problems modelled with \cemph{all-different} constraints that stem from CSPLib~\cite{gewa99a}.
%Note that \emph{permutation} is a special case of \cemph{all-different}, i.e., the number of variables is equal to the number of all their possible values.
%An encoding of \emph{permutation} extends (4) by integrity constraints
%\[
%\leftarrow \dneg e(v_1, i), \dots, \dneg e(v_n, i)
%\]
%or (8) by cardinality rules
%\[
%\leftarrow d-u+l\ \{ \dneg r(v_1, l, u), \dots, \dneg r(v_n, l, u) \}
%\]
%where $1 \leq l \leq u \leq k$~\cite{drwa10a}.
%This can increase propagation.
%
Experiments consider different options in our translation-based approach to constraint answer set solving. We denote the support encoding by~\encsup, the bound encoding by~\encbou, and the range encoding by~\encran. To explore the impact of small Hall intervals, we also tried \encbouh{k} and \encranh{k}, an encoding with only those cardinality rules (\ref{range:all}) for which $u-l+1 \leq k$. The consistency achieved by \encbouh{k} and \encranh{k} may be weaker than bound and range consistency, respectively, when $k < n$.
We also include the hybrid CASP systems \systemname{clingcon} (0.1.2), and \systemname{ezcsp} (1.6.9) in our empirical analysis.
While \systemname{clingcon} extends the ASP system \systemname{clingo} (2.0.2) with the CP solver \systemname{gecode} (2.2.0), \systemname{ezcsp} combines the grounder \systemname{gringo} (2.0.3) and ASP solver \systemname{clasp} (1.3.0) with \systemname{sicstus} (4.0.8) as CP solver.
(Note that the system \systemname{clingo} combines the grounder \systemname{gringo} and ASP solver \systemname{clasp} in a monolithic way.) To provide a representative comparison with \systemname{clingcon} and \systemname{ezcsp}, we have applied \systemname{clingo} (2.0.3) to the encodings in our translation-based approach.
To compare the performance of constraint answer set solvers against traditional CP, we also report results of \systemname{gecode} (3.2.0). Its heuristic for variable selection was set to a smallest domain as in \systemname{clingcon}.
All experiments were run on a 2.00~GHz PC under Linux. We report results in seconds, where each run was limited to 600 s time and 1 GB RAM.

\paragraph{Pigeon Hole Problems}
The famous \emph{pigeon hole problem} is to show that it is not possible to assign $n$ pigeons to $n-1$ holes if each pigeon must be assigned a distinct hole. As can be seen from the results shown in Table \ref{tab:php}, our bound and range encodings perform significantly faster compared to weaker encodings and the other options using filtering algorithms for the \cemph{all-different} constraint that achieve arc consistency on its binary decomposition. However, as can be expected on such problems, detecting large Hall intervals is essential.
\begin{table}[t]
\centering
\setlength{\tabcolsep}{1.8pt}
\begin{tabular}{cccccccccc} \hline\hline
$n$ & \encsup & \encbouh{1} & \encbouh{3} & \encbou & \encranh{3} & \encran & \systemname{ezcsp} & \systemname{clingcon} & \systemname{gecode} \\ \hline
10 & 5.4 & 0.7 & 0.1 & \textbf{0.0} & 0.2 & \textbf{0.0} & 1.8 & 1.4 & 0.9 \\
11 & 46.5 & 3.5 & 1.0 & \textbf{0.0} & 1.9 & \textbf{0.0} & 16.7 & 15.2 & 9.0 \\
12 & 105.0 & 14.8 & 3.9 & \textbf{0.0} & 2.6 & 0.1 & 183.9 & 172.5 & 104.1 \\
13 & --- & 91.4 & 25.4 & 0.1 & 30.4 & \textbf{0.0} & --- & --- & --- \\
14 & --- & --- & 125.0 & \textbf{0.0} & 196.9 & 0.1 & --- & --- & --- \\
15 & --- & --- & --- & \textbf{0.1} & --- & \textbf{0.1} & --- & --- & --- \\
%16 & --- & --- & --- & \textbf{0.1} & --- & \textbf{0.1} & --- & --- & --- \\
\hline\hline
\end{tabular}
\caption{Runtime results in seconds for pigeon hole problems. \label{tab:php}}
\vspace{-1.5\baselineskip}
\end{table}

\paragraph{Quasigroup Completion}
A \emph{quasigroup} is an algebraic structure over $n$ elements and can be represented by an $n \times n$-multiplication table such that each element in the structure occurs exactly once in each row and each column of the table. The \emph{quasigroup completion problem} is to show whether a partially filled table can be completed to a multiplication table of a quasigroup.
We have included models for \systemname{gecode} that enforce bound and domain consistency on \cemph{all-different}, denoted \systemname{gecode}$_{B}$ and \systemname{gecode}$_{D}$, respectively, in our experiments.
Table \ref{tab:qcp} gives the runtime for solving QCP of size $n=20$. The left-most column gives the ratio of preassigned entries.
The results demonstrate phase transition behaviour in the systems \systemname{ezcsp}, \systemname{clingcon}, \systemname{gecode}, and \systemname{gecode}$_{B}$, while our ASP encodings and \systemname{gecode}$_{D}$ (not shown) solve all problems within seconds.
We conclude that learning constraint interdependencies as in our approach (using CDNL) is sufficient to tackle quasigroup completion, i.e., specialised algorithms that enforce domain consistency are not necessary.
\begin{table}[t]
\centering
\setlength{\tabcolsep}{3pt}
\begin{tabular}{cccccccc} \hline\hline
\% & \encsup & \encbou & \encran & \systemname{ezcsp} & \systemname{clingcon} & \systemname{gecode} & \systemname{gecode}$_{B}$ \\ \hline
10 & \textbf{2.6} & 8.2 & 7.3 &29.6 (7) & 9.7 (4) & 2.2 (4) & 0.5 (1) \\
20 & \textbf{2.4} & 8.0 & 7.2 &21.3 (20) & 6.2 (5) & 5.0 (4) & 0.9 (3) \\
30 & \textbf{2.3} & 7.9 & 7.1 &10.3 (30) & 12.9 (13) & 2.9 (13) & 1.1 (5) \\
35 & \textbf{2.3} & 7.9 & 7.0 &21.6 (24) & 11.2 (17) &14.1 (13) & 6.2 (7) \\
40 & \textbf{2.3} & 7.8 & 6.9 &51.6 (29) & 23.1 (22) &11.7 (20) & 5.7 (9) \\
45 & \textbf{2.3} & 7.8 & 6.8 &36.3 (35) & 14.7 (28) &17.7 (25) & 6.3 (13) \\
50 & \textbf{2.3} & 7.7 & 6.8 &36.1 (50) & 21.2 (37) &25.1 (32) & 6.3 (18) \\
55 & \textbf{2.3} & 7.6 & 6.7 &61.4 (51) & 24.4 (44) &19.6 (41) &30.9 (29) \\
60 & \textbf{2.2} & 7.5 & 6.6 &60.2 (63) & 31.4 (56) &36.0 (51) &27.2 (35) \\
70 & \textbf{2.2} & 7.1 & 6.0 &70.0 (66) & 30.2 (50) &28.0 (45) &17.0 (27) \\
80 & \textbf{2.1} & 6.7 & 5.5 &16.2 (18) & 4.2 (18) &17.2 (13) & 7.0 (7) \\
90 & 2.1 & 6.7 & 5.5 & \textbf{1.4} & 2.6 (1) & 0.4 (1) & 3.2 \\
\hline\hline
\end{tabular}
\caption{Average times over 100 runs on quasigroup completion problems. Timeouts, if any, are given in parenthesis. \label{tab:qcp}}
\vspace{-1.5\baselineskip}
\end{table}

\paragraph{Quasigroup Existence}
The \emph{quasigroup existence problem} is to determine the existence of certain interesting classes of quasigroups with some additional properties (\cite{fuslbe93a}).
The properties are represented by axioms \#1 -- \#7 in the direct encoding. In \systemname{ezcsp} and \systemname{gecode}, we additionally use constructive disjunction. Their logic programming equivalent are integrity constraints, exploited in the options $S$, $B_k$, $R_k$ and \systemname{clingcon}. As for \systemname{ezcsp} and \systemname{clingcon} on benchmark classes \#1 to \#4, our resuls presented in Table \ref{tab:qep} suggest that both constructive disjunction and integrity constraints have a similar behaviour.
However, our encodings benefit again from learning constraint interdependencies, resulting in runtimes that outperform all other systems including \systemname{gecode} on the hardest problems.
\begin{table}[t]
\centering
\setlength{\tabcolsep}{1.8pt}
\begin{tabular}{cccccccccc} \hline\hline
\# & $n$ & \encsup & \encbouh{1} & \encbouh{3} & \encbou & \encran & \systemname{ezcsp} & \systemname{clingcon} & \systemname{gecode} \\ \hline
1 & $7$ & 1.7 & 1.7 & 1.7 & 1.7 & 1.6 & 65.0 & 189.8 & \textbf{0.6} \\
1 & $8$ & 19.0& 5.9 & \textbf{4.7} &19.8 & \textbf{4.7} & --- & --- & --- \\
1 & $9$ & --- & \textbf{139.4} & 152.0& 234.6 & 466.9 & --- & --- & --- \\ 
2 & $7$ & 1.7 & 1.7 & 1.7 & 1.8 & 1.8 & 46.1 & 1.5 & \textbf{1.2} \\
2 & $8$ & 46.6& \textbf{9.6} & 10.6 &37.7 & 14.8 & --- & --- & --- \\
2 & $9$ & --- & 246.0 & \textbf{55.7}& 88.3 & 213.4 & --- & --- & --- \\ 
3 & $7$ & 0.2 & 0.2 & 0.2 & 0.3 & 0.3 & 3.2 & 1.0 & \textbf{0.0} \\
3 & $8$ & 0.4 & 0.4 & 0.5 & 0.5 & 0.5 & 4.3 & 9.0 & \textbf{0.2} \\
3 & $9$ &10.2 &\textbf{7.4} & 9.5 &16.5 & 12.8 & --- & --- & 18.2 \\ 
4 & $7$ & 0.2 & 0.2 & 0.2 & 0.3 & 0.3 & 2.8 & 0.7 & \textbf{0.1} \\
4 & $8$ & 0.5 & 0.6 & 0.7 & 0.9 & 0.7 &27.9 &36.8 & \textbf{0.3} \\
4 & $9$ & 1.3 & \textbf{1.0} & 2.1 &3.0 & 0.9 &442.1&288.8& 3.7 \\ 
%& $8$ & 0.4 & 0.4 & 0.4 & 0.5 & 0.4 & 6.9 & 5.3 & \textbf{0.0} \\
%& $9$ & 0.7 & 0.8 & 0.8 & 0.9 & 0.8 &249.2& --- & \textbf{0.0} \\
5 & $10$& 1.6 & 1.5 & 1.6 & 1.9 & 1.6 & --- & --- & \textbf{0.2} \\
5 & $11$& 2.1 & 2.2 & 2.4 & 3.4 & 2.4 & --- & --- & \textbf{0.8} \\
5 & $12$&27.0 &\textbf{6.2}& 9.1 &12.4 & 10.4 & --- & --- & 16.4 \\ 
%& $8$ & 0.4 & 0.4 & 0.5 & 0.5 & 0.4 & 0.8 & --- & \textbf{0.0} \\
%& $9$ & 0.7 & 0.7 & 0.8 & 0.9 & 0.8 & 1.2 & --- & \textbf{0.0} \\
6 & $10$& 1.2 & 1.4 & 1.5 & 1.8 & 1.5 &10.5 & --- & \textbf{0.1} \\
6 & $11$& 2.7 & 2.8 & 4.0 & 4.2 & 4.8 &125.5& --- & \textbf{1.2} \\
6 & $12$&32.0 &\textbf{12.9}& 25.6 &36.4 & 50.6 & --- & --- & 24.6 \\ 
7 & $8$ & 0.4 & 0.4 & 0.4 & 0.6 & 0.5 & 1.1 & --- & \textbf{0.1} \\
7 & $9$ & \textbf{0.7} & 1.0 & 1.2 & 1.7 & 1.4 & 9.1 & --- & 0.9 \\
7 & $10$ & 6.7 & \textbf{3.2} & 5.2 & 8.0 & 4.6 & --- & --- & 22.0 \\ \hline\hline
\end{tabular}
\caption{Results in seconds for quasigroup existence.\label{tab:qep}}
\end{table}

\paragraph{Graceful Graphs}
A labelling of the nodes in a graph~$(V,E)$ is \emph{graceful} if it assigns a unique label from the integers in $[0,|E|]$ such that, when each edge is labelled with the distance between its nodes' labels, the resulting edge labels are all different.
The \emph{graceful graph problem} is to determine the existence of such a labelling.
We use auxiliary variables for edge labels. Their relation to node labels is represented in the direct encoding which weakens the overall consistency.
Table~\ref{tab:ggp} shows our results for \emph{double wheel graphs}, i.e., graphs composed of two copies of a cycle with $n$~nodes, each connected to a central hub.
Our encodings compete with \systemname{ezcsp} and outperform the other systems, whilst the support encoding performs better than bound and range encodings. We observe some variability in the results for $B_k$ and $R_k$, e.g., for $n=8$ the options $B_1$ and $B$ solve the problem within the time limit but $B_3$ does not, although $B_3$ contains $B_1$. We explain this variability by the lookback-based branching heuristic used by \systemname{clingo} being misled by the extra variables introduced in $B_k$ and $R_k$. This is inherent to a growing size of the encoding.
\begin{table}[t]
\centering
\setlength{\tabcolsep}{2.5pt}
\begin{tabular}{ccccccccc} \hline\hline
$n$ & \encsup & \encbouh{1} & \encbouh{3} & \encbou & \encran & \systemname{ezcsp} & \systemname{clingcon} & \systemname{gecode} \\ \hline
%$3$ & 11.4 &  3.8 &  5.7 &  8.7 & 10.4 & 6.5 & 66.9 & \textbf{1.8} \\
$4$ &  1.3 &  2.0 &  1.5 &  3.2 &  2.5 & 0.6 & \textbf{0.1} & \textbf{0.1} \\
$5$ &  4.5 &  5.0 &  4.5 & 13.5 & 31.4 & 1.0 & 2.0 & \textbf{0.1} \\
$6$ &  7.2 & 11.0 & 17.6 & 47.7 &110.2 & \textbf{1.2} & --- & 7.2 \\
$7$ & 23.8 & 28.3 & 67.9 &227.9 &432.9 & \textbf{18.0} & --- & --- \\
$8$ & 48.4 & 68.4 & ---  &207.8 &356.8 & \textbf{4.3} & --- & --- \\
$9$ & \textbf{82.8} &106.5 &200.4 &486.6 &227.4 & 390.5 & --- & --- \\ \hline\hline
\end{tabular}
\caption{Results in seconds for graceful graph problems. \label{tab:ggp}}
\end{table}

\section{Related Work}
Most previous work integrates CP techniques into ASP to avoid huge ground instantiations given through logic programs with first-order variables over large domains. An ASP system was extended in~\cite{baboge05a,mege08a,megezh08a} such that it does not require full grounding, since variables and limitations on their domains can be handled in the CP solver.
A similar approach presented in~\cite{padoporo09a} employs the CP solver to compute also the answer sets.
Although these hybrid strategies potentially eliminate the bottleneck that is inherent to the translation-based approach, they view ASP and CP solvers as blackboxes which do not match the performance of state-of-the-art SMT solvers. In particular, they do not make use of conflict-driven learning and back-jumping techniques.
This gap was closed by the approach taken in~\cite{geossc09a} following the one by~SMT solvers in letting the ASP solver deal with the propositional structure of the logic program, while a CP solver addresses the constraints. Apart from extending the unit-propagation of an ASP solver through constraint propagation, it deals with the elaboration of reasons for atoms derived by constraint propagation within conflict resolution. The elaboration of conflict information from constraint propagators, however, is limited since constraint propagators lack support for this feature (they would have to keep an implication graph to record reasons for each propagation step). Hence, the conflict resolution process cannot exploit constraint interdependencies.
A different hybrid approach to solving CASP is presented in~\cite{balduccini09a}, where an answer set of a logic program with constraint atoms encodes a desired CSP which, in turn, is handled by a CP system. A more general framework using multiple declarative paradigms to specify CSP is proposed in \cite{jaoijani09a}. Either approach, however, restricts communication between different solver types in order to compute solutions to the whole CASP model, e.g., they also do not incorporate conflict-driven learning and back-jumping techniques.

In a translation-based approach, all parts of the model are mapped into a single constraint language for which highly efficient off-the-shelf solvers are available. Hence, related work has mostly focussed on the translation of constraints to SAT (cf. \cite{wa00,gent02}). Translation into ASP, however, can be more general than translation into SAT: Every nogood can be syntactically represented by a clause, but other ASP constructs are also possible, such as cardinality and weight constraints~\cite{siniso02a}. 
ASP was put forward as a novel paradigm for modelling and solving CSP in~\cite{niemela99a}, where straightforward encodings to represent generic constraints via either allowed or forbidden combination of values has been presented.
Preliminary work on translating CASP into ASP was conducted in~\cite{gehiscth09a}, but they did not consider what level of consistency was achieved by their translation.

Decompositions of \cemph{all-different} into simple arithmetic constraints such that bound and range consistency can be achieved were proposed in~\cite{bekanaquwa09a}. There is no polynomial-sized decomposition that achieves domain consistency~\cite{bekanawa09a}.

\section{Conclusions}
We have shown that constraint answer set programming is a promising approach to representing and solving combinatorial problems that naturally merges CP and ASP, while preserving the advantages of both paradigms.
We have presented a translation-based approach to constraint answer set solving. In particular, we have proposed various generic ASP encodings for constraints on finite domains such that the unit-propagation of an ASP solver achieves a certain type of local consistency.
We have formulated our techniques as a preprocessor that can be applied to existing ASP systems without changing their source code. This allows for programmers to select the solver that best fit their needs.
An empirical evaluation of the computational impact on benchmarks from CP has shown our approach outperforming CP and hybrid CASP systems on most instances. As a key advantage we have identified that CDNL exploits constraint interdependencies which can improve propagation between constraints.

Future work concerns the combination of our translation-based approach with a hybrid CASP system centred around lazy nogood generation (cf. lazy clause generation in \cite{ohstco09a}) to combine the advantages of either approach. We will also explore the different choices that arise from this combination.

\paragraph{Acknowledgements}
%Part of this work was performed when Christian Drescher was supported by the European Master's Program in Computational Logic (EMCL).
NICTA is funded by the Department of Broadband, Communications and the Digital Economy, and the Australian Research Council.

\end{document}